\documentclass[11pt]{article}

\usepackage[margin=1in]{geometry}

\usepackage{amsthm,amsmath,amssymb,mathtools}
\usepackage{enumitem,graphicx,color,url}
\usepackage{authblk,setspace,booktabs}
\usepackage[numbers,sort&compress]{natbib}

\author[1]{Naeem Shahabi Sani\thanks{Email: shahabi@ou.edu}}
\author[1]{Shayan Shafaei}
\author[1]{Luyuan Yang}
\author[1]{Dean Hougen}
\author[1]{Chao Lan}
\affil[1]{School of Computer Science, 
University of Oklahoma}
\date{}

\begin{document}

\title{Efficient Low-Rank Matrix Recovery based on\\ 
Randomized Least Squares}

\maketitle

\begin{abstract}
Low-rank matrix recovery is a fundamental task. 
Two common recovery techniques are alternating 
least squares (ALS) and nuclear norm minimization, 
which require iterative optimization of the 
recovered matrix which can be computationally 
expensive. In this paper, 
we propose a substantially faster alternative 
based on randomized least squares (RLS), 
which requires no iterative 
optimization. Experimental results 
on six real-world datasets demonstrate 
its efficacy, where RLS performs similarly 
to these counterparts but is often at least 
10 times faster. 
\end{abstract}

\section{Introduction}

Low-rank matrix recovery is a fundamental 
task with broad applications \cite{davenport2016overview}. 
Two common recovery techniques are alternating least squares (ALS) \cite{jain2013low} and nuclear norm minimization (NuNo) \cite{recht2010guaranteed}. 
The former iteratively refines two low-rank factors 
that are used for recovery, and the latter iteratively refines a recovered low-rank matrix using algorithms like singular value thresholding \cite{cai2010singular}. 
They both rely on iterative optimization of the recovered matrix,  which can be computationally expensive. 

In this paper, we introduce a substantially 
faster alternative based on randomized 
least squares (RLS), which does not need 
iterative optimization. It is similar 
to ALS, but only optimizes one factor in 
one shot while randomly constructing the 
other factor. 
RLS has been studied for matrix factorization \cite{zhou2011godec,persson2023randomized,saha2023matrix}, 
but not for matrix recovery to our best knowledge. 
Unlike the factorization task, the 
recovery task begins with an incomplete 
target matrix which hinders the 
application of RLS (for it takes a 
complete matrix as input). 
We overcome this challenge by a heuristic, 
which efficiently imputes the target matrix 
and then applies RLS to refine the result. 
For convenience, we also refer to the 
resulting recovery technique as RLS. 

Our experimental results on six real-world 
datasets demonstrate the efficacy of RLS, 
showing that it performs similarly to ALS, NuNo 
and their efficient variants but often 
runs at least 10 times faster. We also 
validate the heuristic by showing that 
RLS outperforms the initial imputation step.

\section{Randomized Least Squares 
for Matrix Recovery}
\label{sec:rls}

\textbf{Notation.}
 For matrix $M$, we use $M_{ij}$ to denote its
entry in row $i$ and column $j$, $M_{i:}$ to 
denote its $i_{th}$ row, $M_{:j}$ to denote 
its $j_{th}$ column, and $\|M\|_F$ to denote 
its Frobenius norm.

\vspace{10pt} 
\noindent \textbf{Setting.}
Let $X \in \mathbb{R}^{n \times m}$ be 
an incomplete matrix (to recover) and 
$\Omega$ be the 
index set of its observed entries. 
Let $U \in \mathbb{R}^{n \times k}$ 
and $V \in \mathbb{R}^{k \times m}$ 
be two low-rank factors, where 
$k < \min(m, n)$. The ALS 
technique iteratively optimizes 
$(U,V)$ so that their product 
can well approximate $X$, i.e., 
\begin{equation}
X \approx U V.        
\end{equation}

\vspace{5pt} 
\noindent \textbf{Technique.}
We propose to optimize the factors 
in one shot based on RLS. The
Standard RLS method requires $X$ to be complete, 
whereas ours is incomplete. We overcome 
this by a heuristic that relies 
on imputation. Specifically, the proposed 
recovery technique has three major steps. 

\vspace{10pt} 
\noindent \textit{Step 1}. Impute the missing 
values of $X$ with 
values i.i.d. sampled from $N(\hat{\mu}, 
\hat{\sigma}^2)$, 
where $(\hat{\mu},\hat{\sigma})$ are the 
mean and standard deviation of the observed 
entries in $X$. Let $\tilde{X}$ be the 
imputed matrix. 

\vspace{10pt} 
\noindent \textit{Step 2}. 
Generate one factor 
\begin{equation}
\label{eq:setU}
U = \tilde{X} W,     
\end{equation}
where $W \in \mathbb{R}^{m\times k}$ is a 
random matrix with i.i.d. $N(0,1/k)$ entries. 

\vspace{10pt} 
\noindent \textit{Step 3}. 
Optimize the other 
factor by solving 
\begin{equation}
\label{eq:rlsrecoversol}
\min_{V} \sum_{(i,j) \in \Omega}
\left( X_{ij} - U_{i:} V_{:j} \right)^2 + \lambda \|V\|_F^2,
\end{equation}
where $\lambda$ is a regularization coefficient. 

\vspace{10pt} 
We can verify that the solution to 
(\ref{eq:rlsrecoversol}) is, for 
each column $j = 1, \ldots, m$, 
\begin{equation}
\label{eq:solV}
V_{:j} =  (U_{\Omega_j}^T U_{\Omega_j} 
+ \lambda I)^{-1} U_{\Omega_j}^T X_{\Omega_j},  
\end{equation}
where $I$ is identity, 
$\Omega_j = \{i': (i',j) \in \Omega\}$ 
is the row index set of observed entries 
at column $j$  and
\begin{equation}
U_{\Omega_j} = \left[
\begin{matrix}
U_{i_1 :} \\ \vdots 
\\ U_{i_{|\Omega_j|} :} 
\end{matrix} \right] \in 
\mathbb{R}^{|\Omega_j| \times k} 
\quad \text{and}\quad 
\hfill X_{\Omega_j} = \left[
\begin{matrix}
X_{i_1 j} \\ \vdots 
\\ X_{i_{|\Omega_j|} j} 
\end{matrix} \right] \in 
\mathbb{R}^{|\Omega_j|},
\end{equation}
where $i_1, \ldots, i_{|\Omega_j|}$ are 
the elements of $\Omega_j$. 
After obtaining $(U,V)$, $X$ is recovered as 
\begin{equation}
\label{eq:recovery}
\hat{X} = \text{round}(U V),  
\end{equation}
where the round function maps 
recovered values to their 
nearest legitimate values, e.g., integers. 

\subsection{Time Complexity Analysis}
\label{sec:complexity}

The time complexity of RLS is 
$O(mnk + mk^3 + |\Omega|k^2)$. 
This includes $O(mn)$ for imputing 
$X$ (Step 1), $O(mnk)$ for computing 
$U$ (Step 2), $O(k^3 + |\Omega_j| k^2)$ 
for computing each $V_{:j}$ (Step 3) 
so the total time is $O(mk^3 + |\Omega| k^2)$, 
and $O(mnk)$ for recovering $X$ based on 
(\ref{eq:recovery}). 

The time complexity of ALS is 
$O(Tnk^3 + Tmk^3 + T|\Omega|k^2 + mnk)$,
where $T$ is the number of rounds 
for factor update. This includes 
$O(mnk)$ for computing $UV$ and 
$O(nk^3 + mk^3 + |\Omega|k^2)$ 
time for one round of factor updates 
(which consists of $O(mk^3 + |\Omega| k^2)$ 
for updating $V$ and 
$O(nk^3 + |\Omega| k^2)$ 
for updating $U$). Compared to RLS, 
ALS is less efficient primarily 
due to the repeated updates. 
If $T = 1$ and $k^2 = O(m)$, the two 
recovery techniques 
have the same time complexity. (In this 
case, however, our experimental results 
show that ALS performs much worse than 
RLS; see Table~\ref{tab:comparison}).

We also compare RLS with CSMC 
\cite{krajewska2026randomized}, 
a recent and efficient 
recovery technique. It randomly 
samples $c$ ($>k$) columns of $X$ 
to form a submatrix 
$X_c \in \mathbb{R}^{n \times c}$, 
then recovers it 
using any technique, 
and finally reconstructs a complete 
$X$ from the recovered $X_c$ through 
least squares regression. 

For a fairer comparison, assume 
RLS is used to recover $X_c$. 
Then the time of CSMC is 
$O(cnk + ck^3 + |\Omega|k^2 
+ |\Omega|c^2 + mc^3 + mnc) 
= O(mnc + k^3c + |\Omega| c^2 + mc^3)$. 
In the left big O, the 
first three terms are for running 
RLS on $X_c$, and the last 
three terms are for reconstructing $X$ 
from $X_c$ -- this includes regressing 
the observed entries in $X$ from the 
recovered $X_c$ columns which takes $O(|\Omega|c^2 
+ mc^3)$ time\footnote{To be specific, 
CSMC learns one linear model to predict 
each column $X_{:j}$ from an $|\Omega_j| \times c$ submatrix of $X_c$ which contains the 
rows of $X_c$ indexed by $\Omega_j$. The model 
is learned by the regularized least squares 
technique.}, and recovering 
$X$ by multiplying $X_c$ with 
the regression coefficient matrix  
which takes $O(mnc)$ time. 
If we view RLS as a regression 
process from $U$ to $X$, then 
it corresponds to the regression 
process from $X_c$ to $X$. 
Based on this view, CSMC is less efficient 
than RLS for two reasons: its 
$X_c$ is no smaller than $U$ 
($c > k$), and it takes an extra 
 time for recovering $X_c$. 

Even if we recover $X_c$ 
using a faster technique 
NuNoR \cite{feng2018faster}, 
CSMC remains less efficient. 
In this case, the time of CSMC becomes
$O(Tnck + |\Omega|c^2 + mc^3 + mnc)$, 
where $O(Tnck)$ is the approximate 
time for NuNoR to recover $X_c$ and 
the rest is for regressing $X$ from $X_c$. 
Compared to the time of RLS, this 
remains less efficient because 
$c > k$ and it incurs an extra $O(Tnck)$ term. 

In \cite{krajewska2026randomized} and our 
experiments, $c$ is set to a fixed fraction of 
$m$ i.e., $c = \Theta(m)$. Then, the above 
CSMC time becomes $O(Tmnk + |\Omega|m^2 + 
m^4 + m^2n)$, which grows faster than that 
of RLS or ALS as the target matrix size increases.

\section{Experiment}

We experimented on six datasets:
MovieLens-100K and MovieLens-1M \cite{harper2015movielens},
FilmTrust \cite{massa2007trust}, Jester \cite{goldberg2001eigentaste},
BookCrossing \cite{ziegler2005diversification}, 
and Yahoo R3
\cite{yahoo}. On each set, we used 
the predefined training and test splits 
whenever available; otherwise, 80\% of the observed entries were randomly selected for training 
and the rest were used for testing. 
Following \cite{tegetmeier2024artificial}, 
the BookCrossing matrix was downsampled to $2000\times2500$. Table~\ref{tab:dataset_dimensions}
summarizes the statistics of these datasets. 

\begin{table}[t]
\centering
\renewcommand{\arraystretch}{1.1}
\setlength{\tabcolsep}{6pt}
\normalsize
\begin{tabular}{lrr}
\toprule
Data set & Rows (users) & Columns (items) \\
\midrule
MovieLens-100K & 943    & 1,682 \\
MovieLens-1M   & 6,040  & 3,706 \\
FilmTrust      & 1,508  & 2,071 \\
Jester         & 24,983 & 100   \\
BookCrossing   & 2,000  & 2,500 \\
Yahoo R3       & 15,400 & 1,000 \\
\bottomrule
\end{tabular}
\caption{Statistics of Datasets}
\label{tab:dataset_dimensions}
\end{table}

We compare RLS with ALS, NuNo, and two 
efficient variants: NuNoR, which applies 
randomized singular value thresholding in NuNo \cite{feng2018faster}, and Columns Selected Matrix Completion (CSMC), which recovers a submatrix 
and uses it to reconstruct the target matrix \cite{krajewska2026randomized}. For CSMC, 
NuNoR is used to recover the sampled submatrix. 

The main hyperparameters are the regularization coefficients for ALS and RLS, and the column sampling 
rate for CSMC. We optimized the former 
based on sensitivity analysis, and fixed the latter 
to 0.1 based on experience (i.e., $c/m = 0.1$). 
All baseline methods are iterative; each stops when 
the change of mean squared difference between
consecutive recovered matrices falls below 
a tolerance or a maximum number of
iterations is reached.
We evaluate recovery performance based on two metrics: classification error
(i.e., the fraction of mis-recovered entries after rounding) and recovery time. All reported results are
averaged over 20 runs with different random initializations.
The source code is available 
at\footnote{\url{https://github.com/naeemshb/Randomized-matrix-recovery}}.

\subsection{Performance Comparison}
\label{sec:overall}

Experimental results are shown in
Table~\ref{tab:performance_comparison}. 
RLS is often 10$\sim$100 times faster 
than other methods. In terms 
of recovery accuracy, RLS is better 
than ALS on BookCrossing, MovieLens-100K 
and Yahoo, while falls behind on other 
sets ($\sim 3\%$ lower). 

NuNo and NuNoR often achieve 
the best accuracies, but they are computationally 
inefficient, e.g., NuNoR is 10 times slower 
than RLS and NuNo is often 100 times slower. 
While RLS generally sacrifices accuracy 
to obtain these speedups, it is encouraging 
to see that the tradeoff is not always 
hard, e.g., on MovieLens-100K RLS is only $1\%$ 
less accurate than NuNo and NuNoR. 

\begin{figure}[t!]
\centering
\includegraphics[width=.5\linewidth]{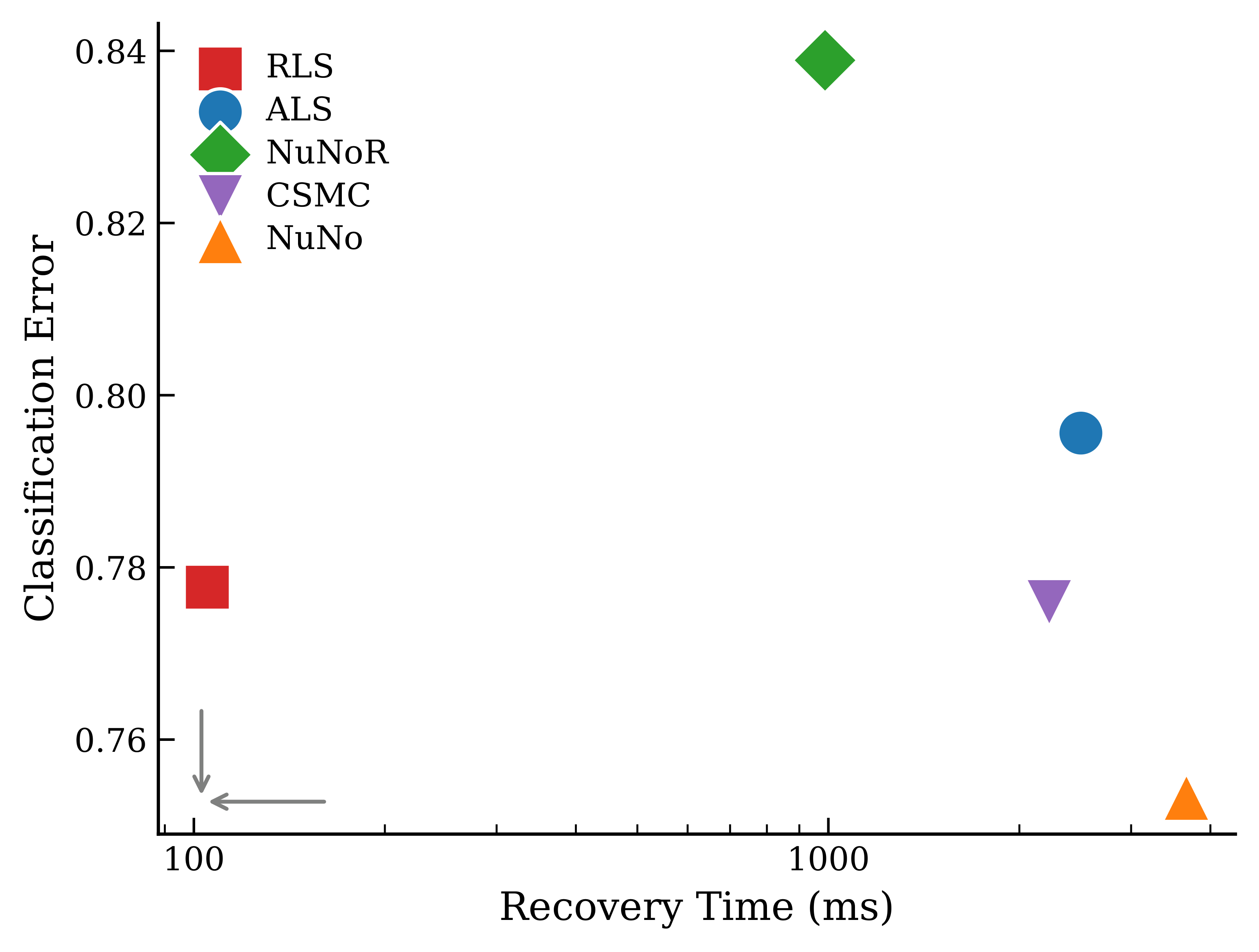}
\vspace{-5pt}
\caption{Accuracy-Time Tradeoff on BookCrossing. 
Lower-left is better.}
\label{fig:Pareto}
\end{figure}

Figure~\ref{fig:Pareto} visualizes 
the accuracy and time of all techniques 
on BookCrossing ($k=20$). 
Compared to other methods, RLS is in 
the lower-left corner, suggesting 
a more efficient tradeoff.

\begin{table*}[t]
\centering
\normalsize 
\renewcommand{\arraystretch}{1.2}
\setlength{\tabcolsep}{2pt}
\begin{tabular}{l r@{\,$\pm$\,}l r@{\,$\pm$\,}l @{\hspace{12pt}} r@{\,$\pm$\,}l r@{\,$\pm$\,}l}
\toprule
& \multicolumn{4}{c}{MovieLens-100K ($k=15$)} & \multicolumn{4}{c}{BookCrossing ($k=20$)} \\
\cmidrule(r{8pt}){2-5} \cmidrule(l){6-9}
Method & \multicolumn{2}{c}{Error} & \multicolumn{2}{c}{Time (ms)} & \multicolumn{2}{c}{Error} & \multicolumn{2}{c}{Time (ms)} \\
\midrule
CSMC  & 0.7951 & 2e-3 & 1055 & 16.78 & 0.7759 & 3.4e-3 & 1957 & 79.76 \\
NuNo  & 0.7940 & 5e-4 & 5557 & 133.32 & \textbf{0.7533} & 8e-4 & 3557 & 11.09 \\
NuNoR & \textbf{0.7854} & 7e-4 & 500 & 3.45 & 0.8389 & 1.7e-3 & 988 & 5.21 \\
ALS   & 0.7974 & 4e-4 & 445 & 18.34 & 0.7956 & 2.2e-3 & 2500 & 151.82 \\
RLS   & 0.7947 & 8.8e-3 & \textbf{46} & 1.62 & 0.7777 & 4.7e-3 & \textbf{95} & 0.54 \\
\midrule & \multicolumn{4}{c}{FilmTrust ($k=20$)} 
& \multicolumn{4}{c}{MovieLens-1M ($k=17$)} \\
\cmidrule(r{8pt}){2-5} \cmidrule(l){6-9}
Method & \multicolumn{2}{c}{Error} & \multicolumn{2}{c}{Time (ms)} & \multicolumn{2}{c}{Error} & \multicolumn{2}{c}{Time (ms)} \\
\midrule
CSMC  & 0.7747 & 8.5e-3 & 697 & 22.54 & 0.7848 & 1.1e-3 & 8651 & 27.0 \\
NuNo  & 0.7751 & 1.4e-3 & 3217 & 0.12 & 0.7851 & 3e-4 & 304829 & 23.4 \\
NuNoR & \textbf{0.7493} & 1.4e-3 & 781 & 5.45 & 0.7746 & 2e-4 & 7247 & 107.0 \\
ALS   & 0.7753 & 1.7e-3 & 890 & 266.52 & \textbf{0.7705} & 4e-4 & 10317 & 127.0 \\
RLS   & 0.8041 & 5.9e-3 & \textbf{64} & 2.02 & 0.7967 & 1.5e-3 & \textbf{522} & 27.0 \\
\midrule
& \multicolumn{4}{c}{Jester ($k=30$)} & \multicolumn{4}{c}{Yahoo R3 ($k=30$)} \\
\cmidrule(r{8pt}){2-5} \cmidrule(l){6-9}
Method & \multicolumn{2}{c}{Error} & \multicolumn{2}{c}{Time (ms)} & \multicolumn{2}{c}{Error} & \multicolumn{2}{c}{Time (ms)} \\
\midrule
CSMC  & 0.8077 & 1.0e-3 & 953 & 45.02 & 0.7945 
& 3.2e-3 & 2048 & 23.02 \\
NuNo  & 0.8097 & 1.0e-4 & 1058 & 7.69 & 0.8514 & 
2.0e-4 & 30485 & 34.43 \\
NuNoR & \textbf{0.7881} & 5.0e-4 & 1110 & 51.80 & 0.8261 & 3.0e-4 & 4945 & 2.98 \\
ALS   & 0.7978 & 7.0e-4 & 1780 & 11.84 & 0.5998 & 2.8e-3 & 2891 & 60.70 \\
RLS   & 0.8391 &  2.5e-3 & \textbf{109} & 2.12 & \textbf{0.5967} & 3.1e-2 & \textbf{263} & 9.49 \\
\bottomrule
\end{tabular}
\caption{Recovery Performance 
(mean $\pm$ standard deviation)}
\label{tab:performance_comparison}
\end{table*}

We also generated synthetic data matrices
to evaluate how the recovery time scales 
with data size. These matrices have sizes 
ranging from $500\times400$ to $7500\times6000$ 
with ranks fixed to $k = 20$ (which is also 
used by the evaluated methods). 
In each matrix, $10\%$ of entries 
were randomly sampled for 
training and the rest for testing. 
Hyperparameters were selected once using a $1000\times800$ validation matrix and then 
fixed across all matrix sizes. 
ALS and CSMC were each run for exactly 10 iterations. 
All results were averaged over five random trials, where 
each trial uses a different random seed to generate the 
matrix, to sample the observed entries, and to initialize 
the methods.
Recovery time is measured as the wall-clock time of the 
recovery routine, as on the real-world data.

\begin{table}[t]
\centering
\normalsize
\renewcommand{\arraystretch}{1.1}
\setlength{\tabcolsep}{6pt}
\begin{tabular}{lrrrrr}
\toprule
& \multicolumn{3}{c}{Time (ms)} & \multicolumn{2}{c}{Speedup of RLS} \\
\cmidrule(lr){2-4} \cmidrule(l){5-6}
Size & RLS & CSMC & ALS & vs.\ CSMC & vs.\ ALS \\
\midrule
$500{\times}400$   & 10   & 24    & 159   & 2.4$\times$  & 15.8$\times$ \\
$1000{\times}800$  & 30   & 92    & 403   & 3.1$\times$  & 13.5$\times$ \\
$2000{\times}1600$ & 94   & 349   & 1120  & 3.7$\times$  & 11.9$\times$ \\
$3000{\times}2400$ & 191  & 1103  & 1982  & 5.8$\times$  & 10.4$\times$ \\
$5000{\times}4000$ & 524  & 4890  & 4981  & 9.3$\times$  & 9.5$\times$ \\
$7500{\times}6000$ & 1078 & 19224 & 10175 & 17.8$\times$ & 9.4$\times$ \\
\bottomrule
\end{tabular}
\caption{Recovery Time on Synthetic Data.}
\label{tab:runtime_scalability}
\end{table}

Table \ref{tab:runtime_scalability} shows the 
recovery time. As the matrix size increases, 
the time of all methods increases. 
RLS remains faster than ALS and leads CSMC by increasingly
large margins. This is consistent with our 
time complexity analysis in Section \ref{sec:complexity}: the time of RLS and ALS scale more slowly 
than that of CSMC as the target matrix 
size increases. 

\subsection{Validation of the Heuristic}
\label{sec:imputation}

RLS is built on a heuristic that first 
imputes the target matrix and then uses 
it to generate a low-rank factor. 
It is natural to ask whether the 
imputation alone is sufficient. 
In this section, we address this question 
by comparing the performance of 
RLS and its initial imputation. 

\begin{figure*}[t]
    \centering

    \begin{minipage}[t]{0.32\textwidth}
        \centering
        \includegraphics[width=\linewidth]{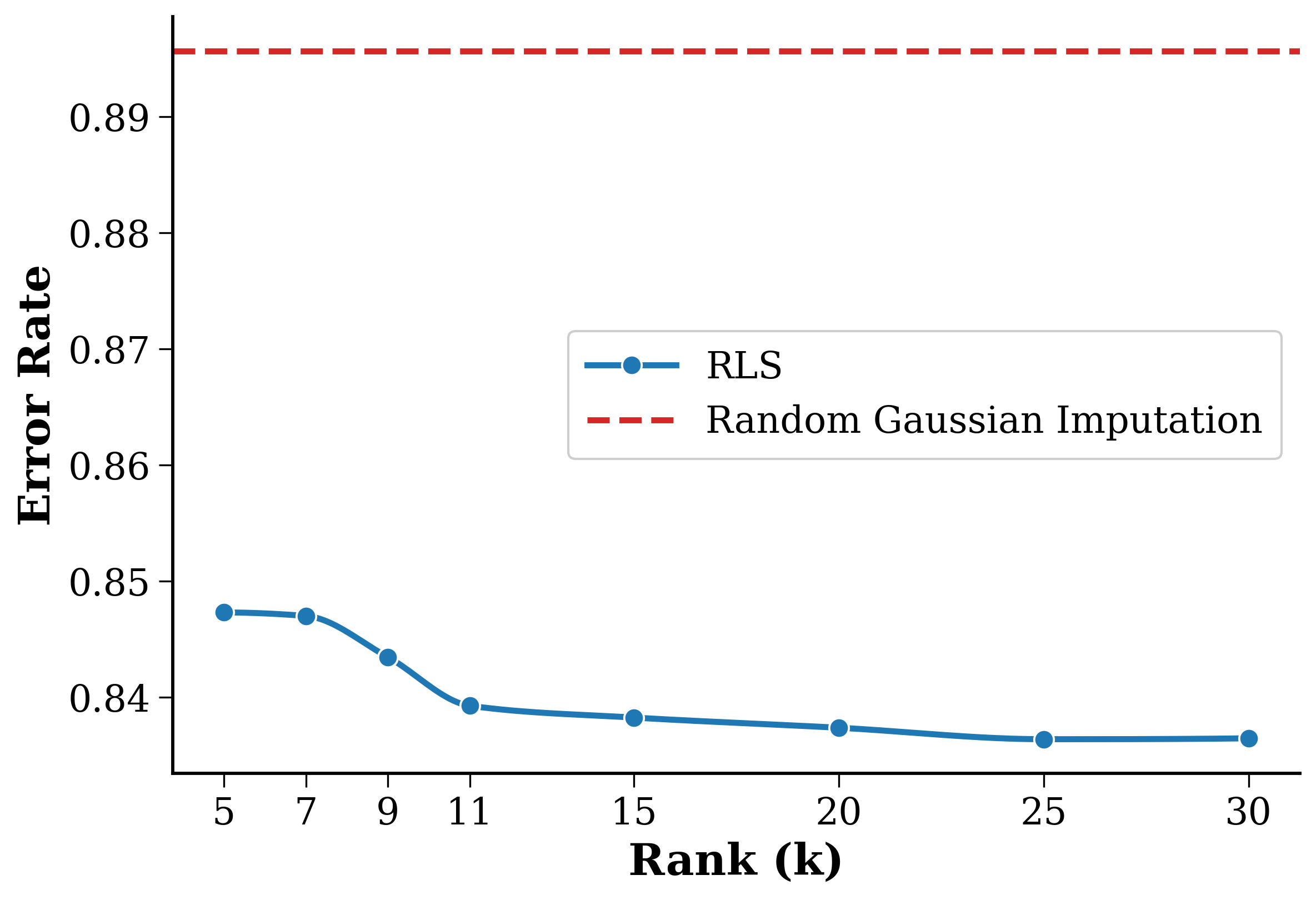}
        \small (a) Gaussian Imputation 
        \end{minipage}
        \hfill 
    \begin{minipage}[t]{0.32\textwidth}
        \centering
        \includegraphics[width=\linewidth]{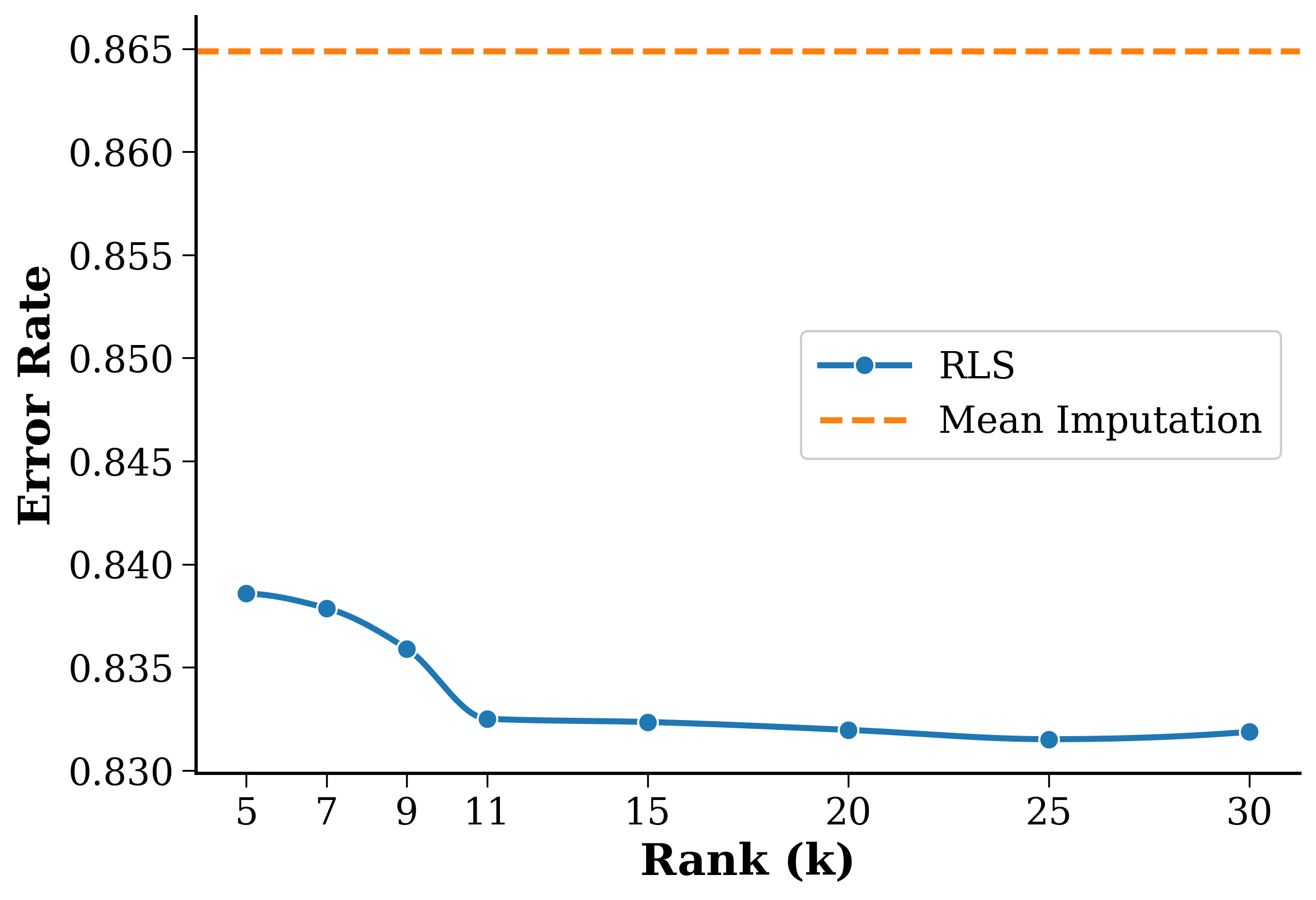}
        \small (b) Mean Imputation
    \end{minipage}
    \hfill
    \begin{minipage}[t]{0.32\textwidth}
        \centering
        \includegraphics[width=\linewidth]{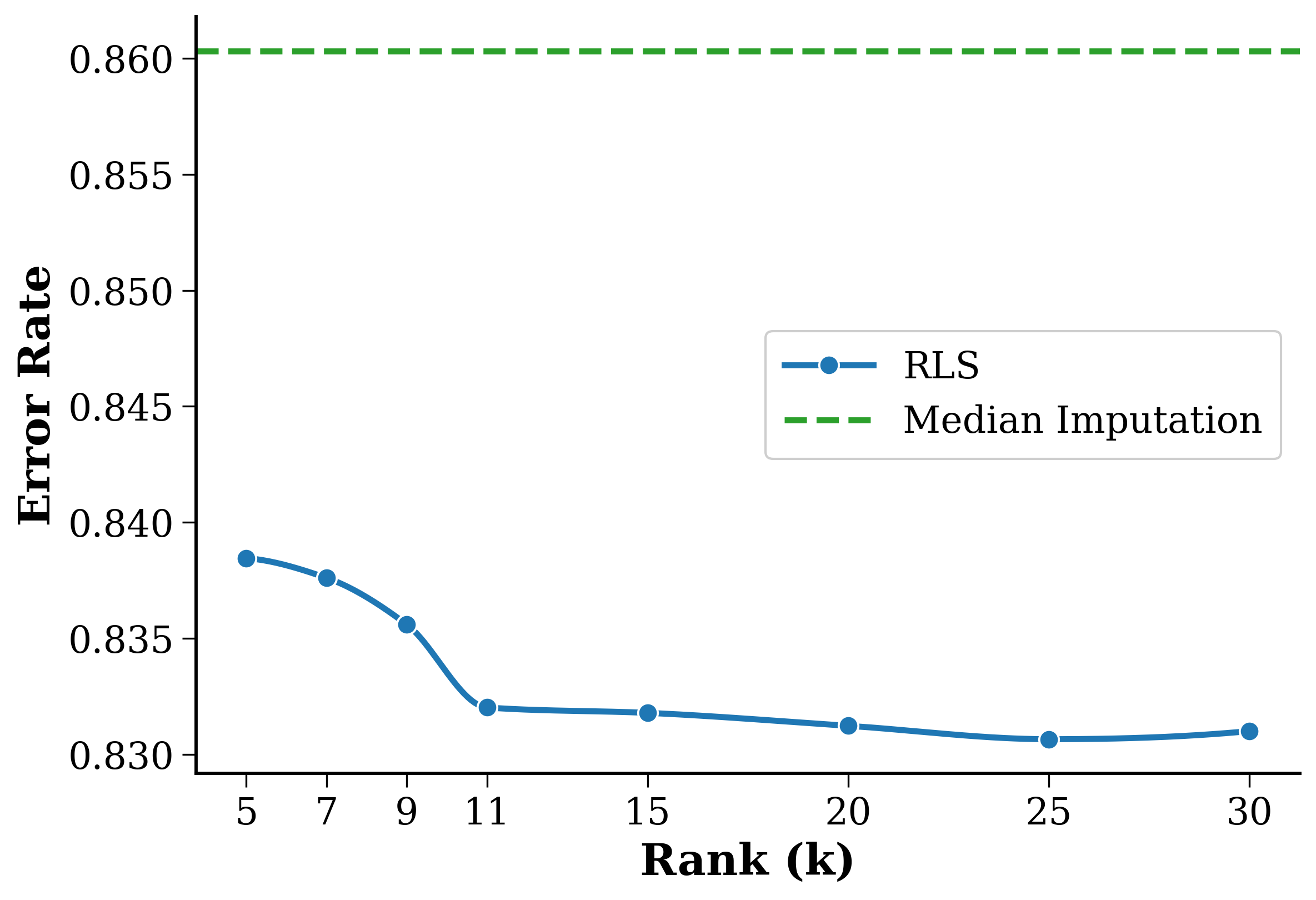}
        \small (c) Median Imputation
    \end{minipage}
    \caption{Classification Errors on Jester}
    \label{fig:imputation_comparison}
\end{figure*}

Figure \ref{fig:imputation_comparison} 
shows the results on Jester. RLS 
consistently outperforms the corresponding 
imputation method across different choices 
of $k$, which validates its effectiveness. 
In each figure, RLS applies the same imputation 
as the competing method.

\begin{table}[t]
\centering
\normalsize
\renewcommand{\arraystretch}{1.1}
\setlength{\tabcolsep}{6pt}
\begin{tabular}{c|c|c|c}
\toprule
Training Data & {Gaussian Imputation} 
& {Mean Imputation} & {Median Imputation} \\
\midrule
100\% &  6.59\% & 3.66\% &  3.18\% \\
80\%  & 5.87\% & 2.97\% &  2.49\% \\
60\%  &  5.18\% &  2.31\% &  1.83\% \\
40\%  &  4.52\% &  1.66\% &  1.19\% \\
20\%  &  3.68\% &  0.71\% &  0.30\% \\
\bottomrule
\end{tabular}
\caption{Relative Error Reduction on Jester.}
\label{tab:jester_sparsity}
\end{table}

Table~\ref{tab:jester_sparsity} shows 
the relative error reduction, defined as
$(E_{\mathrm{imp}}-E_{\mathrm{RLS}})
/ E_{\mathrm{imp}}$, where $E_{\mathrm{imp}}$ is 
the recovery error based on imputation alone 
and $E_{\mathrm{RLS}}$ is the recovery error 
of RLS. In this table, 
$k = 30$ and the training data size varies 
from $100\%$ to $20\%$ of the original 
training data size. We see that 
RLS outperforms the imputation method 
across different cases.

\subsection{Diagnostic Comparison with One-Round ALS}
\label{sec:onepass}

As discussed earlier, ALS has the same time  
complexity as RLS when it is limited to one round 
of factor update. In this section, we 
evaluate the resulting recovery performance. 

Results on FilmTrust ($k=20$) are shown in 
Table \ref{tab:comparison}. Results of RLS 
are slightly different from those in Table 
\ref{tab:performance_comparison} due to 
different random trials.  
Compared to RLS, 
ALS (limited to one round of factor update) 
has a similar recovery time but much higher 
error. This further justifies the value of 
RLS, that not any naive one-round recovery 
update can maintain recovery performance.

\begin{table}[t]
\vspace{20pt}
\centering
\normalsize
\renewcommand{\arraystretch}{1.2}
\setlength{\tabcolsep}{6pt}
\begin{tabular}{lcc}
\toprule
Method & Error & Time (ms) \\ \midrule
ALS & 0.8811 $\pm$ 0.0087 & 81.62 $\pm$ 0.78 \\
RLS & 0.8032 $\pm$ 0.0053 
& 72.43 $\pm$ 0.84 \\
\bottomrule
\end{tabular}
\caption{ALS (one round) versus RLS on FilmTrust.}
\label{tab:comparison}
\end{table}





\section{Related Work}
\label{sec:related}
Low-rank matrix completion has been widely studied 
from the perspective of convex optimization. 
Cand\`es and Recht~\cite{candes2009exact} show that,
under suitable incoherence and sampling conditions, a low-rank
matrix can be recovered exactly from a subset of its entries by
minimizing the nuclear norm subject to agreement with the observed
entries. Cand\`es and Plan~\cite{candes2010matrix} study recovery from
noisy observations, with recovery error bounded in terms of the
noise level. Cai et al.~\cite{cai2010singular} develop an iterative
singular value thresholding algorithm that applies SVD-based
shrinkage at each iteration.
Another line of work uses low-rank factors to model 
observed ratings, as described by Koren et al.~\cite{koren2009matrix}.
ALS alternately updates these factors through least squares;
Jain et al.~\cite{jain2013low} establish convergence guarantees
under suitable conditions. 

A few prior studies employed randomized algorithms to accelerate matrix recovery. For example, NuNoR 
\cite{feng2018faster} accelerates NuNo by replacing the SVD step in its singular value thresholding procedure with
randomized SVD, but the procedure remains iterative. 
A recent study \cite{krajewska2026randomized}
proposes Columns Selected Matrix Completion (CSMC), 
which selects a subset of columns from the target matrix, completes the resulting 
submatrix using a standard recovery technique (e.g., 
ALS or NuNo), and reconstructs the full matrix from 
the recovered submatrix. While recovering a smaller 
matrix is indeed faster, the applied recovery algorithm 
often remains iterative. 

Recent work has continued to improve iterative recovery 
methods. Gu et al.~\cite{gu2023low} propose a robust alternating
minimization framework for low-rank matrix completion that achieves
nearly linear runtime by replacing exact regression solves with efficient
sketching-based approximations, while Ma and Fattahi~\cite{ma2024convergence}
show that vanilla gradient descent with small initialization provably
converges to the ground truth for symmetric positive-semidefinite matrix
completion without requiring explicit regularization, even in the
over-parameterized setting. Zhang et al.~\cite{zhang2024orthogonal}
propose an orthogonal subspace exploration approach together with a
smooth non-convex surrogate for matrix completion; their optimization
consists of closed-form updates and empirically converges within dozens
of iterations. Bauch et al.~\cite{bauch2021rank} propose R2RILS, which
uses an over-parameterized rank-$2r$ representation and solves a
least-squares problem at each iteration to jointly update the row and
column subspaces.  These recovery techniques remain 
iterative.

\section{Conclusion}
\label{sec:conclusion}

In this paper, we propose an efficient matrix 
recovery technique based on randomized least squares 
(RLS). Unlike popular recovery techniques that 
require iterative optimization, RLS optimizes the 
recovery model in one round. Through experiments 
on six real-world datasets, we show RLS runs substantially 
faster while maintaining similar recovery accuracy 
in most cases. 

\bibliographystyle{plain}
\bibliography{reference}

\end{document}